\documentclass[sigconf]{acmart}

\usepackage{booktabs} 
\usepackage{amsmath,amssymb,amsfonts}
\usepackage{algorithmic}
\usepackage[linesnumbered,ruled]{algorithm2e}
\usepackage{graphicx}
\usepackage{textcomp}
\usepackage{xcolor}
\usepackage{etoolbox,siunitx}
\usepackage{subcaption}

\usepackage{geometry}

\SetAlFnt{\scriptsize}
\SetAlCapFnt{\scriptsize}
\SetAlCapNameFnt{\scriptsize}

\makeatletter
\renewcommand{\algocf@caption@boxruled}{%
	\hrule
	\hbox to \hsize{%
		\vrule\hskip-0.4pt
		\vbox{   
			\vskip\interspacetitleboxruled%
			\unhbox\algocf@capbox\hfill
			\vskip\interspacetitleboxruled
		}%
		\hskip-0.4pt\vrule%
	}\nointerlineskip%
}%
\makeatother

\setcopyright{rightsretained}

\acmDOI{10.1145/nnnnnnn.nnnnnnn}

\acmISBN{978-x-xxxx-xxxx-x/YY/MM}

\acmConference[GECCO '19]{the Genetic and Evolutionary Computation Conference 2019}{July 13--17, 2019}{Prague, Czech Republic}
\acmYear{2019}
\copyrightyear{2019}

\acmPrice{15.00}

\begin{document}
\title{Evolving Deep Neural Networks by Multi-objective Particle Swarm Optimization for Image Classification}

\author{Bin Wang, Yanan Sun, Bing Xue and Mengjie Zhang}
\affiliation{%
	\department{School of Engineering and Computer Science}
	\institution{Victoria University of Wellington}
	\streetaddress{PO Box 600}
	\city{Wellington} 
	\country{New Zealand}
	\postcode{6140}
}
\email{{bin.wang, yanan.sun, bing.xue, mengjie.zhang}@ecs.vuw.ac.nz}

\renewcommand{\shortauthors}{B. Wang et al.}
\renewcommand{\shorttitle}{Evolving Deep NNs by Multi-objective PSO for Image Classification}

\begin{abstract}
In recent years, convolutional neural networks (CNNs) have become deeper in order to achieve better classification accuracy in image classification. However, it is difficult to deploy the state-of-the-art deep CNNs for industrial use due to the difficulty of manually fine-tuning the hyperparameters and the trade-off between classification accuracy and computational cost. This paper proposes a novel multi-objective optimization method for evolving state-of-the-art deep CNNs in real-life applications, which automatically evolves the non-dominant solutions at the Pareto front. Three major contributions are made: Firstly, a new encoding strategy is designed to encode one of the best state-of-the-art CNNs; With the classification accuracy and the number of floating point operations as the two objectives, a multi-objective particle swarm optimization method is developed to evolve the non-dominant solutions; Last but not least, a new infrastructure is designed to boost the experiments by concurrently running the experiments on multiple GPUs across multiple machines, and a Python library is developed and released to manage the infrastructure. The experimental results demonstrate that the non-dominant solutions found by the proposed method form a clear Pareto front, and the proposed infrastructure is able to almost linearly reduce the running time. 
\end{abstract}

%
%



\copyrightyear{2019}
\acmYear{2019}
\setcopyright{acmcopyright}
\acmConference[GECCO '19]{Genetic and Evolutionary Computation Conference}{July 13--17, 2019}{Prague, Czech Republic}
\acmBooktitle{Genetic and Evolutionary Computation Conference (GECCO '19), July 13--17, 2019, Prague, Czech Republic}
\acmPrice{15.00}
\acmDOI{10.1145/3321707.3321735}
\acmISBN{978-1-4503-6111-8/19/07}

\maketitle

\section{Introduction}

Image classification has been attracting more and more interest both from the academic and industrial researchers due to the exponential growth of images in terms of both the number and the resolution, and the meaningful information extracted from images. Convolutional Neural Networks (CNNs) have been investigated to solve the image classification tasks. Although CNNs were introduced more than 20 years ago, the classification accuracy has been significantly improved on hard problems in recent years because the rapid development of hardware capacity makes it possible to train very deep CNNs. A couple of years ago, VGG \cite{VGG_Simonyan}, which was deemed as a very deep CNN, only had 19 layers, but the recently-proposed ResNet \cite{ResNet_He} and DenseNet \cite{DenseNet_Huang} were capable of effectively training CNNs of more than 100 layers, which dramatically reduced the classification error rate. 

\begin{figure*}[t]
	\centering
	\includegraphics[width=0.9\textwidth]{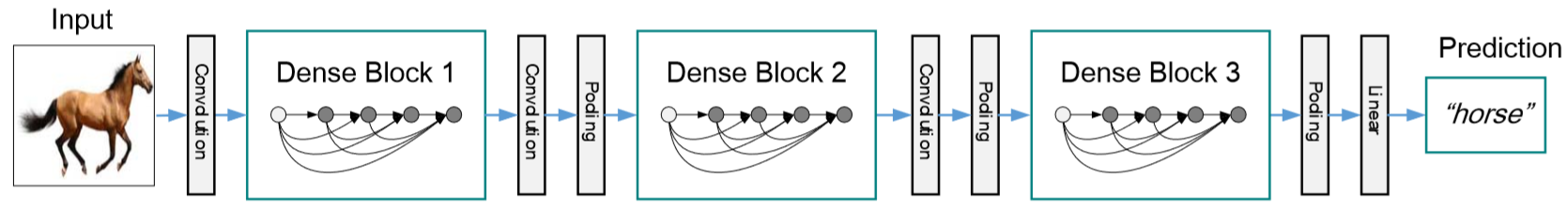}
	\caption{DenseNet architecture (Image taken from \cite{DenseNet_Huang})}
	\label{fig:densenet_architecture}
\end{figure*}

However, it is hard to deploy CNNs in real-life applications mainly because of the following two reasons: 

\begin{itemize}
	\item The state-of-the-art CNNs are designed by experts, and tuning the hyperparameters of CNNs to fit the dataset in a specific application is complex and time-consuming;
	\item A trade-off between the classification accuracy and inference latency needs to be made, which is hard to be decided by application developers.
\end{itemize}

Taking DenseNet as an example, although several DenseNet architectures are evaluated in its paper, there are two obstacles for applying DenseNet in real-life applications: Firstly, the hyperparameters may not be optimized, and for different tasks, the optimal model is not fixed, so before integrating DenseNet into applications, the hyperparameters have to be fine-tuned, which can be very complicated; Secondly, since an optimal DenseNet for a specific task may be extremely deep, the inference latency can be too long in some real-life applications such as web applications or mobile applications given limited hardware resource. This means that the classification accuracy may need to be comprised by reducing the complexity of DenseNet in order to reduce the inference latency to an acceptable amount of time. 

In recent years, neural architecture search (NAS) \cite{NAS_Elsken} \cite{NAS_Zoph}, which automatically search for optimal models by optimizing the hyperparameters of neural networks, has been drawing the attention of interested researchers. However, computational cost is very high for most of the methods \cite{EvolutionCNN_Moore} \cite{EvolutionNAS_Real}. Some research work \cite{PSOCNN_Wang} \cite{DECNN_Wang} \cite{GeneticCNN_Xie} has been done to successfully reduce the computational cost. One of the widely-used methods to reduce the computational cost in most of the researches is to save the evaluation time of individuals in the evolutionary process by using a small number of the training epochs and training the model on partial datasets, which may bring some uncertainty to the search space. In this paper, we focus on solving the difficulty of applying the-state-of-the-art CNNs for industrial use. In order to minimise the search space, a specific type of CNN architecture can be chosen, so a smaller search space is formed based on special domain knowledge of the experts. In addition, most of the researches in NAS focus on improving the classification accuracy, but inference latency is critical for real-life applications.

\subsection{Goals}

The overall goal of this paper is to propose a multi-objective particle swarm optimization (MOPSO) method to balance the trade-off between the classification accuracy and the inference latency, which is named MOCNN. MOCNN automatically tunes the hyperparameters of CNNs and deploys the trained model for better industrial use. To be more specific, an MOPSO method will be developed to search for a Pareto front of models. Therefore, industrial users can obtain the most suitable model for their specific problem based on their image classification task and the target devices. The specific objectives of this work are listed below:

\begin{enumerate}
	\item As DenseNet achieved the competitive classification accuracy comparing the state-of-the-art methods, in order to reduce the search space, this work will focus on optimizing the hyperparameters of Dense blocks, such as the number of dense blocks, the growth rate of each dense block, and the number of layers of each dense block. An encoding strategy will be proposed to encode the dense blocks; 
	\item There are two major factors - classification accuracy and computational cost, which are decisive to the performance of the neural network. This work will develop an MOPSO application to optimize the hyperparameters of dense blocks by jointly considering the classification accuracy and the computational cost. The specific two objectives are classification accuracy and FLOPs (floating point operations) where FLOPs can reflect the computational cost of both training and inference; 
	\item Completely training a CNN, which is required by objective evaluations, is dramatically slower than applying the operations of evolutionary computation (EC) algorithms, which becomes the bottleneck of the computational cost of the whole evolutionary process. In order to speed up the experiment, a server-client GPU infrastructure will be designed and a python library will be developed to concurrently train a batch of CNNs across multi-GPUs on multi-servers. 
\end{enumerate}

\section{Background}\label{sec:Background}


\subsection{DenseNet}\label{sec:Background_DenseNet}

A DenseNet is composed of several dense blocks, which are connected by a convolutional layer followed by a pooling layer, and before the first dense block, the input is filtered by a convolutional layer. An example of a DenseNet comprising three dense blocks is outlined in Fig. \ref{fig:densenet_architecture}. Apart from the dense blocks, the hyperparameters of the other layers are fixed. The hyperparameters for the convolutional layer before the first block are problem-specific based on the image size in order to reduce the image size of the input feature maps passed to the first block; while the hyperparameters of the layers between blocks are problem-agnostic, which are a 1$\times$1 convolutional layer and a 2$\times$2 average pooling layer. However, the hyperparameters of dense blocks vary depending on specific image classification tasks, which are the number of layers in the dense block and the \textit{growth rate} of the dense block. The \textit{growth rate} is the number of output feature maps for each convolutional layer in the dense block. The output $x_{l}$ is calculated according to Formula (\ref{eq:DenseNet_ouput}), where $[x_{0}, x_{1}, ..., x_{l-1}]$ refers to the concatenation of the feature maps obtained from layer 0, 1, ..., $l-1$, and $H_{l}$ represents a composite function of three consecutive operations, which are batch normalization (BN) \cite{BN_Ioffe}, a rectified linear unit (ReLU) \cite{ReLU_Glorot} and $3\times3$ convolution (Conv).

\begin{equation}\label{eq:DenseNet_ouput}
x_{l} = H_{l}([x_{0}, x_{1}, ..., x_{l-1}])
\end{equation}

\subsection{OMOPSO}\label{sec:BackgroundOMOPSO}

\begin{algorithm}[ht]
	\caption{OMOPSO}
	\label{alg:omopso}
	\begin{algorithmic}[1]
		\renewcommand{\algorithmicrequire}{\textbf{Input:}}
		\renewcommand{\algorithmicensure}{\textbf{Output:}}
		\STATE $P, A \leftarrow$ Initialize swarm, initialize empty $\epsilon-archive$;
		\STATE $g \leftarrow$ Set the current generation $g$ to 0;
		\STATE $L \leftarrow$ Select leaders from $P$;
		\STATE Send $L$ to $A$;
		\STATE \textit{crowding}($L$)
		\WHILE{$g<g_{max}$}
		\FOR{$particle$ \textbf{in} $P$}
		\STATE Select leader, updating, mutation and evaluation
		\STATE Update \textit{pbest}
		\ENDFOR
		\STATE $L \leftarrow$ Update leaders
		\STATE Send $L$ to $A$;
		\STATE \textit{crowding}($L$)
		\STATE $g \leftarrow g+1$
		\ENDWHILE
		\STATE Report results in $A$
	\end{algorithmic}
\end{algorithm}

OMOPSO \cite{OMOPSO_Sierra} is a multi-objective optimization approach based on Pareto dominance, which selects the leaders using a crowding factor. The pseudo code of OMOPSO is written in Algorithm \ref{alg:omopso}. There are a few items in the algorithm that need to be pointed out. First of all, there are two archives used by the algorithm: the first archive stores the current leaders that are used for performing the updating, and the other one carries the final solutions. The leaders are selected based on the crowding values of the leaders, while the final solutions are the non-dominant solutions according to $\epsilon$-Pareto dominance \cite{EpsilonPareto_Laumanns}. In addition, when the maximum number of leaders is exceeded, the crowding factor \cite{NSGA2_Deb} \cite{NSPSO_Li} is used to filter out the leaders based on the crowding values of the leaders in order to keep the number of leaders within the maximum number limit. Thirdly, for each particle, when selecting a leader for the updating of OMOPSO, the binary tournament based on the crowding value is applied. Finally, the particles are divided into three parts of equal size, and three mutation schemes are applied on the three parts, respectively. The first part has no mutation at all, the second part has uniform mutation, and the third part has non-uniform mutation. 


\section{The Proposed Method}\label{sec:ProposedAlgorithm}


\subsection{Algorithm Overview}\label{sec:ProposedAlgorithm_Overview}

The framework of the proposed MOCNN has three steps. The first step is to initialise the population described in Section \ref{sec:PopulationInitialization} based on the proposed particle encoding strategy illustrated in Section \ref{sec:ParticleEncodingStrategy}; At the second step, the multi-objective PSO algorithm called OMOPSO \cite{OMOPSO_Sierra} is applied to optimize the two objectives, which are the classification accuracy and the FLOPs; Lastly, the non-dominant solutions in the Pareto set are retrieved, from which the actual user of the CNNs can choose one based on the usage requirements.

\begin{figure}[!t]
	\centering
	\includegraphics[width=0.9\linewidth]{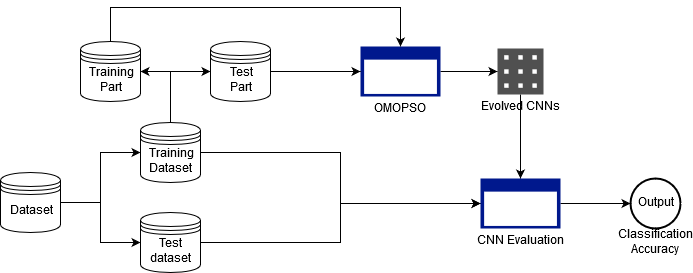}
	\caption{The flowchart of the experimental process}
	\label{fig:expriment_workflow}
\end{figure}

Fig. \ref{fig:expriment_workflow} shows the overall structure of the system. The dataset is split into a training set and a test set, and the training set is further divided into a training part and a test part. The training part and the test part are passed to the proposed OMOPSO method. During the objective evaluation, the training part is used to train the neural network, and the test part is used to obtain the test accuracy of the trained neural network, which is used as the objective of classification accuracy. The proposed OMOPSO method produces non-dominant solutions, which are the optimized CNN architectures. Depending on the trade-off between the classification accuracy and the hardware resource capability, one of the non-dominant solutions can be selected for actual use. The CNN evaluation needs to be fine-tuned for the selected CNN architecture, and the whole training set and the test set are used to obtain the final classification accuracy.

\subsection{Particle Encoding Strategy}\label{sec:ParticleEncodingStrategy}

\begin{figure}[ht]
	\centering
	\includegraphics[width=0.9\linewidth]{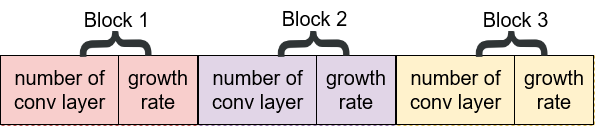}
	\caption{An example of a particle vector}
	\label{fig:MOCNNEncoding}
\end{figure}

In DenseNet, the hyperparameters, which need to be optimized, are the number of bocks, the number of layers in each block, and the growth rate of each block. For each of the block, a vector with the length of two can represent the number of layers and the growth rate in the block. Once the number of blocks is defined, the number of layers and the growth rate in each block can be encoded into a vector with the fixed length of 2 $\times$ the number of blocks. Fig. \ref{fig:MOCNNEncoding} shows an example of the vector, which carries the hyperparameters of DenseNets with 3 blocks. 

As it can be observed in the proposed encoding strategy, the number of blocks need to be set up first, which brings a couple of advantages. First of all, since OMOPSO has proven to work effectively on a continuous search space with fixed dimensions, after fixing the number of blocks, the DenseNet hyperparameters are encoded into vectors of a fixed-length, where OMOPSO can be applied straightforward. In addition, when performing the OMOPSO evolutionary operators, the $ith$ block of particles moves to its optimal position in the search space if the number of blocks is fixed. However, if the number of blocks is not fixed, one way to solve the problem is to mix the hyperparameters together to produce a fixed-length particle in order to perform OMOPSO, which may produce a lot of disturbance in the search space by breaking the idea of moving each block towards its optimal position; another way is to only move the matched blocks to their optimal, which slows down the flying process of particles by keeping some blocks in the previous position. Therefore, the simple and effective solution adopted by the proposed encoding strategy is to fix the number of blocks. 

\subsection{Population Initialization}\label{sec:PopulationInitialization}

Before initializing the population, the range of each dimension has to be worked out first based on the effectiveness of the network and the capacity of hardware resource. If the number of layers in a block is too small, e.g. the number of layers is smaller than 2, there will not be any shortcut connections built in the dense block, and a very small number of feature maps, i.e. a too small growth rate, will not produce effective feature maps either. On the other hand, if the number of layers or the growth rate is too big, the hardware resource required to run the experiment will likely exceed the actual capacity of the hardware. The specific range of each dimension of our experiment will be designed and listed in Section \ref{S:ec_parameters}. 

The initial population is randomly generated based on the range of each dimension, whose pseudocode is composed in Algorithm 2. To be more specific, when randomly generating an individual, a random value is generated according to the range of each dimension from the first dimension until the last dimension; By repeating the individual generation process until the population size is satisfied, the whole initial population with a fixed population size will be successfully generated. 

\begin{algorithm}[ht]
	\caption{Population initialization}
	\label{alg:pop_init}
	\begin{algorithmic}[1]
		\renewcommand{\algorithmicrequire}{\textbf{Input:}}
		\renewcommand{\algorithmicensure}{\textbf{Output:}}
		\REQUIRE particle dimension $d$, population size $p_{s}$, a list of dimension value range $r$;
		\STATE $P \leftarrow$ Empty population set;
		\STATE $i \leftarrow$ 0;
		\WHILE{$i<p_{s}$}
		\STATE $ind \leftarrow$ Empty particle;
		\WHILE{$j<d$}
		\STATE $ind[j] \leftarrow$ Generate a random number within the range $r[j]$;
		\ENDWHILE
		\STATE $P \leftarrow$ Append $ind$ to $P$;
		\ENDWHILE
	\end{algorithmic}
\end{algorithm}

\subsection{Objective Evaluation}\label{sec:FitnessEvaluation}

As the proposed MOCNN simultaneously optimizes the classification accuracy and the FLOPs, in the objective evaluation of MOCNN, both of them are calculated and returned as the objectives of the individual shown in Algorithm \ref{alg:fitness_evaluation}. When obtaining the classification accuracy, before training the individual representing a DenseNet with its specific hyperparameters, the training dataset is divided into two parts, which are the training part and the test part, and then the individual is trained on the training part and evaluated on the test part using a back propagation algorithm with an adaptive learning rate called Adam optimization \cite{Adam_Kingma} with the default settings, which are $\alpha=0.001, \beta_{1}=0.9, \beta_{2}=0.999, and\,\epsilon=10^{-8}$. The optimization target of the proposed MOCNN is to maximize the classification accuracy; In regard to the computational cost, the FLOPs is calculated for the individual, which is used as the second objective, and the proposed MOCNN will try to minimize the number of FLOPs. 

\begin{algorithm}[ht]
	\caption{objective evaluation}
	\label{alg:fitness_evaluation}
	\begin{algorithmic}[1]
		\renewcommand{\algorithmicrequire}{\textbf{Input:}}
		\renewcommand{\algorithmicensure}{\textbf{Output:}}
		\newcommand{\algorithmicbreak}{\textbf{break}}
		\newcommand{\BREAK}{\STATE \algorithmicbreak}
		\REQUIRE individual $ind$, maximum epochs $e_{max}$, accuracy list of trained CNNs $acc_{saved}$;
		\IF{$ind \textbf{ in } acc_{saved}$}
		\STATE $acc_{best} \leftarrow$ retrieve the accuracy of $ind$ from $acc_{saved}$;
		\ELSE
		\STATE $acc_{best}, e_{best} , e \leftarrow$ 0, 1, 0;
		\WHILE{$e<e_{max}$}
		\STATE Apply Adam optimization \cite{Adam_Kingma} to train $ind$ on the training part $d_{t}$;
		\STATE $acc_{t} \leftarrow$ evaluate the trained $ind$ on the test part $d_{t}$;
		\IF{$acc_{t} > acc_{best}$}
		\STATE $acc_{best}, e_{best} \leftarrow$ $acc_{t}, e$;
		\ELSIF{$e - e_{best}$ $>$ 10}
		\BREAK
		\ENDIF
		\ENDWHILE
		\STATE $acc_{saved} \leftarrow$ Append $ind$ and $acc_{best}$ to $acc_{saved}$;
		\ENDIF
		\STATE $flops \leftarrow $ calculated FLOPs of $ind$;
		\STATE $ind \leftarrow$ update the accuracy and FLOPs of $ind$ by $acc_{best}$ and $flops$;
	\end{algorithmic}
\end{algorithm}

Since training CNNs takes much longer time than that of calculating FLOPs, a couple of methods have been implemented to reduce the computational cost of getting the classification accuracy. First of all, an early stop criterion of terminating the training process when the accuracy does not increase in the next 10 epochs is adopted to potentially reduce the epochs of the training process, which as a result, decreases the training time. It worked particularly effective to search for CNN architectures because the complexity of different individuals may vary significantly, which may require a various number of epochs to completely train different individuals. For example, as the CNN architecture can be as simple as one or two layers with a very small number of feature maps, the number of epochs needed to train the CNN can be very small; while the CNN architecture can also be as complicated as one containing hundreds of layers with a really large number of feature maps in each layer, so it requires much more epochs to completely train the complicated CNN. Therefore, it is hard to define a fixed number of epochs used by the objective evaluation to train CNNs with various complexities. Instead, the proposed MOCNN sets a maximum number of epochs, which is large enough to fully train the most complicated CNNs in our search space, and utilizes the early-stop criterion to stop the training process at an earlier stage in order to save the computational cost. In addition, since each individual will be evaluated by the objective evaluation in each generation, there may be a large number of CNNs evaluated across the whole evolutionary process, among which there may be individuals representing the same CNN architecture duplicately trained and evaluated. For the purpose to prevent the same CNN from the duplicate training, the classification accuracy obtained for each individual in the objective evaluation is stored in the memory, which is persisted just before the program finishes, and loaded at the beginning of the program. In the objective evaluation, before training the individual, a search for the individual in the stored classification accuracy is performed first, and the training procedure will be executed only when the search result is empty.

Adam optimization \cite{Adam_Kingma} is chosen as the backpropagation algorithm, and the whole training dataset is used to evaluate the CNNs. As to our best knowledge, two other methods of objective evaluation were found being used in the area of using EC method to automatically design CNN architectures. The first method used in \cite{EvolutionCNN_Moore} and \cite{EvolutionNAS_Real} is to use Stochastic Gradient Descent (SGD) \cite{SGD_Bottou} with a scheduled learning rate, e.g. 0.1 as the learning rate before 100 epochs, and the learning rate divided by 10 at the epoch of 150 and 200, respectively. From the settings of SGD for training VGGNet \cite{VGG_Simonyan}, ResNet \cite{ResNet_He} and DenseNet \cite{DenseNet_Huang}, it can be observed that the SGD settings are quite different, which means that a set of SGD settings may be good for a specific type of CNNs, but may not work well for other types of CNNs. Therefore, it is very hard to perform a fair comparison between two various CNNs that need SGDs with different settings to optimize, which results in the preference of a specific set of CNNs in the EC algorithm. The second method is to train the CNN for a small number of epochs used in \cite{DENSER_ASSun} and \cite{PSOCNN_Wang}. It speeds up the training process by restraining the number of training epochs, which relies on the assumption that the CNN architecture with a good performance at the beginning would perform well in the end, but to our best effort, a strong evidence hasn't been found to prove the assumption in either theoretical or empirical study. As a result, the evolutionary process may prefer the CNN architectures that perform well at the beginning without any guarantee of achieving a good classification accuracy in the end, but it is the classification accuracy in the end that should be used to select CNN architectures. Both of these two methods may introduce some bias toward a specific set of CNN architectures. However, by using the Adam optimization to train the CNNs on the whole training dataset, it could mitigate or even eliminate the bias of the aforementioned two methods because the learning rate will be automatically adapted during the training process based on the CNN architecture and the dataset, and the training process will stop until the convergence of the Adam optimization. So, the objective evaluation method in the proposed MOCNN method is expected to be able to reduce the bias.  

\subsection{Infrastructure Used to Boost MOCNN}\label{sec:MocnnInfra}

\begin{figure}[ht]
	\centering
	\includegraphics[width=0.9\linewidth]{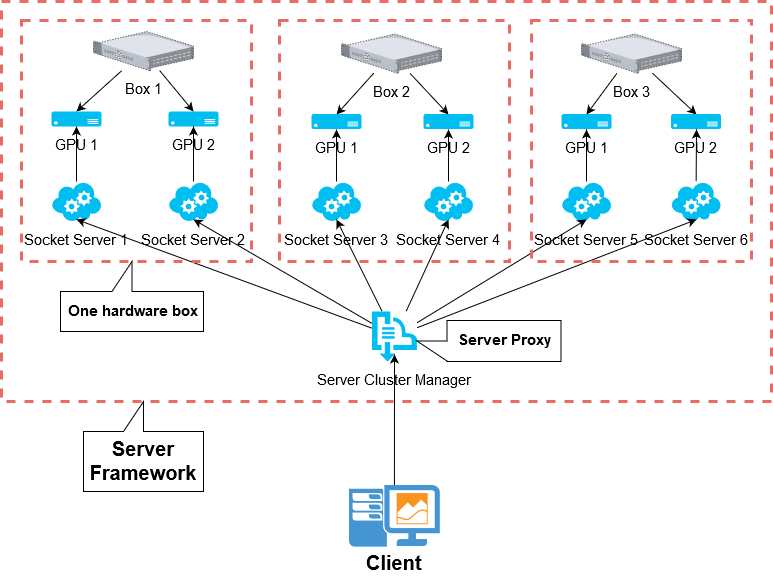}
	\caption{The infrastructure used to boost the experiment}
	\label{fig:experiment_infra}
\end{figure}

As it can be observed, the objective evaluation is the bottleneck for running the proposed MOCNN, and obtaining the classification accuracy by training and evaluating the individual is the bottleneck of the objective evaluation. The common and easy implementation of the objective evaluation would be running it on one GPU card for each individual. One potential method to improve the performance of the training process of CNNs is to leverage multi-GPU functionality provided by the widely-used frameworks to train the CNN on multiple GPUs on one machine to speed up the training process. In order to further reduce the time cost of running the proposed MOCNN, this paper proposes an infrastructure illustrated in Fig. \ref{fig:experiment_infra}, which has the ability to leverage all of the available GPU cards across multiple machines to concurrently perform the objective evaluation for a batch of individuals, and the corresponding python library is developed and published as an open-source python library\footnote{Python library called cudam to manage multi-gpu on multi-servers: \url{https://pypi.org/project/cudam/}}.

There is a cluster of servers running in the infrastructure, which are drawn at the top of the infrastructure diagram in the three boxes. Each of the boxes represents a machine, i.e. a hardware server, containing multiple GPU cards. In the case of the diagram, there are three boxes, which are three machines, where two GPU cards are installed on each box. On each of the GPU card, a socket server is running to listen and handle the request from the client. For example, a CNN that needs to be evaluated may be passed from the client to the server, and the server will train and evaluate the CNN and return the classification accuracy to the client. There are two reasons for using one GPU card as a socket server instead of using the whole box. First of all, as the real infrastructure is likely to comprise hardware boxes with a various number of GPU cards installed, if the whole box is running as a socket server, the capacity of each socket server will be different from each other. While in the proposed MOCNN, the computational cost of training most of the individuals is likely to be similar, and a batch of individuals, the number of which is the same as the number of the socket servers, will be sent to the cluster of servers for objective evaluation. The client expects to collect the batch evaluation results when all of the individuals in the batch have been evaluated to keep the order of the evaluation results of the individuals the same as they are sent, so in order to reduce the idle time of the socket servers, it is better to keep the capacity of the socket servers the same; otherwise, when the client was waiting for the batch evaluation being completed, some socket servers with better capacity may finish much earlier. Secondly, the utilization efficiency of multi-GPU mode depends on the specific framework, and some frameworks cannot reach the optimal usage of multiple GPUs mainly because of the shared resources, which has to be securely shared by multiple threads of the program by locking the shared resource when it is accessed by one thread. However, in the method of using each GPU card as a socket server, it does not have the issue of handling the shared resources, so the efficiency of GPU utilization can be guaranteed. 

In the middle of the infrastructure diagram, there is a server cluster manager, a.k.a., a server proxy, which manages the concurrency of the objective evaluation executed by the cluster of socket servers. Firstly, the proxy server receives all the CNNs that need to be evaluated and store them as a pool of CNNs; Secondly, the server proxy checks the availability of all socket servers in the cluster, and based on the number of available servers, it fetches a batch of unevaluated CNNs, whose number is the same as that of available servers, and distributes each of the CNNs in the batch to one of the available socket servers simultaneously; Thirdly, the proxy server waits to collect the evaluation results for all of the CNNs, which will be attached to the CNNs as the evaluated classification accuracy. By repeating the second and third steps until all of the CNNs in the pool have the classification accuracy attached to them, the server proxy will return the evaluated CNNs back to the client. 

The client is outlined at the bottom of the figure. Any algorithm, which contains an objective evaluation of a number of CNNs, can run as a client in order to leverage the utilization of multiple GPU cards on multiple machines. As the server proxy and the cluster of servers have handled most of the concurrent operations, the usage of the client is really straightforward, which just needs to pass all of the CNNs to the server proxy at once, and wait for the response from the server proxy without any additional management of the concurrent evaluation. 

So far, the details of the infrastructure have been described. However, it would be more understandable to demonstrate how the infrastructure is used to run the proposed MOCNN. Apart from the objective evaluation, the whole EC algorithm runs as a client, which is the main process of the program. However, at the beginning of each generation, all of the individuals in the population will be sent to the server proxy. Once the evaluated individuals are responded from the server proxy, the main program running the EC algorithm will continue. In summary, the proposed method is implemented the same as that of running on a single machine, and the only tweak is to send the individuals to the server proxy for objective evaluation instead of evaluating the CNNs by itself. 

\section{Experiment design}\label{sec:EPDesign}


\subsection{Benchmark Dataset}\label{sec:EPDesign_Benchmark}

\begin{figure}[ht]
	\centering
	\includegraphics[width=0.9\linewidth]{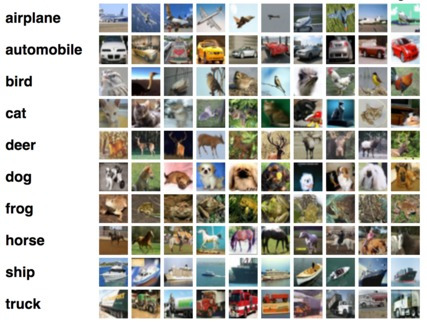}
	\caption{Examples of CIFAR-10 images}
	\label{fig:cifar10_examples}
\end{figure}

Based on the computational cost of the algorithm that needs to be evaluated and the hardware resource to run the experiment, 
the CIFAR-10 dataset will be chosen as the benchmark dataset. It consists of 60,000 colour images with the size of 32$\times$32 in 10 classes, and each class contains 6000 images. The whole dataset is divided into the training dataset of 50,000 images and the test dataset of 10,000 images \cite{CIFAR10_Alex}. Fig. \ref{fig:cifar10_examples} shows the example images from the CIFAR-10 dataset.

\subsection{Parameter settings of the proposed EC methods}\label{S:ec_parameters}

As the proposed method consists of two parts, which are the multi-objective EC algorithm called OMOPSO and the process of training Deep CNNs in the objective evaluation, the parameters listed in Table \ref{table:moccn_parameters} are set according to the conventions of the communities of EC and deep learning with the consideration of the computational cost and complexity of the search space in the proposed MOCNN method. However, several parameters are specific to the proposed MOCNN method, which will be discussed in details in the following paragraphs. 

\begin{table}[ht]
	\renewcommand{\arraystretch}{1.3}
	\caption{Parameter list}
	\label{table:moccn_parameters}
	\centering
	\footnotesize
	\begin{tabular}{|c|c|}
		\hline
		Parameter & Value\\
		\hline
		\multicolumn{2}{|c|}{\textbf{objective evaluation}} \\
		\hline
		initial learning rate & 0.1\\
		\hline
		batch size & 128\\
		\hline
		maximum epochs & 300\\
		\hline
		\multicolumn{2}{|c|}{\textbf{Particle Encoding}} \\
		\hline
		number of blocks & 4\\
		\hline
		the range of growth rate in all four blocks & 8 to 32\\
		\hline
		the range of number of layers in the first block & 4 to 6\\
		\hline
		the range of number of layers in the second block & 4 to 12\\
		\hline
		the range of number of layers in the third block & 4 to 24\\
		\hline
		the range of number of layers in the fourth block & 4 to 16\\
		\hline
		\multicolumn{2}{|c|}{\textbf{OMOPSO}} \\
		\hline
		$\epsilon$ values in the format of [accuracy, FLOPs] & [0.01, 0.05]\\
		\hline
	\end{tabular}
\end{table}

First of all, since the proposed particle encoding strategy is exclusively designed for the proposed MOCNN, the parameters are customized for effectively and efficiently running our MOCNN experiment. As the purpose of this paper is to explore the Pareto front of the multi-objective problem of deep CNNs, this paper is not focusing on setting a new benchmark of the classification accuracy. DenseNet-121, which is the least complex DenseNet reported in the DenseNet paper \cite{DenseNet_Huang}, is chosen as the most complex CNN to be searched by the proposed MOCNN due to our limited memory, computational capacity of our GPU resource and time constraint. Although DenseNet-121 was not the best DenseNet reported in its paper, the classification accuracy was only slightly worse than the more complicated DenseNets, and the computational cost of training DenseNet-121 is quite high, so the least complex DenseNet is set as the maximum complexity given that the training process needs to be performed 400 (20 individuals$\times$20 generations) times in the evolutionary process. As a result, the number of blocks is fixed to 4; 32, which is the growth rate of DenseNet-121, is set as the maximum value of the growth rate; and the maximum number of layers for the first, second, third and fourth block is configured as 6, 12, 24 and 16, respectively, which is the same as that of DenseNet-121. In terms of the lower bound of the parameters, if there are too few layers in a block, the dense connection will not work effectively, and if the growth rate is too small, it will cause the issue of a very limited number of extracted features, which will not provide enough useful features for the classification algorithm. Therefore, 4 and 8 are chosen as the lower bounds of the number of layers in each block and the growth rate, respectively. 

In addition, the maximum epochs used to train the CNNs in objective evaluation is set to 300 based on the number of epochs used to train the most complex CNN in the search space. To be more specific, 100, 200 and 300 epochs were examined for training DenseNet-121 to see whether DenseNet-121 could be fully trained. It turned out training DenseNet-121 for 300 epochs can guarantee the convergence on the CIFAR-10 dataset used as the benchmark dataset in our experiment. 

Furthermore, as the $\epsilon$ value defines the number of non-dominant solutions, which is demonstrated in Section \ref{sec:BackgroundOMOPSO}. A few $\epsilon$ values are investigated for each of the objectives. A smaller value of $\epsilon$ produces fewer non-dominant solutions; while more non-dominant solutions are obtained by increasing the value of $\epsilon$. However, $\epsilon$ value does not affect the evolutionary process of the proposed MOCNN, so the $\epsilon$ value is configured purely based on the number of non-dominant solutions that are preferred to be displayed in the final result, where the actual industrial users of the proposed method can choose the best solution by considering the classification accuracy and the computational cost. 

Finally, the population size and the maximum generation need to be designed for the experiment. 20 and 50 are chosen from the widely-used population sizes based on the convention of the EC community and the high computational cost of our experiment. The reason for running two experiments with different population sizes is to explore how population size will affect the results of the proposed MOCNN method. Due to the time constraint, 400 to 500 evaluations are used, which may take 2 weeks. Therefore, the experiment with 20 individuals will run for 20 generations and the other one with 50 individuals will run for 10 generations. In order to better refer these two experiments, the experiment with 20 individuals and 20 generations is called \textit{EXP-20-20}, and \textit{EXP-50-10} represents the experiment with 50 individuals and 10 generations.

\section{Results and analysis}\label{sec:EPResults}


\subsection{Pareto Optimality Analysis}\label{sec:EPResults_Pareto}

\begin{figure*}[ht]
	\centering
	\begin{subfigure}[b]{0.24\textwidth}
		\centering
		\includegraphics[width=\textwidth]{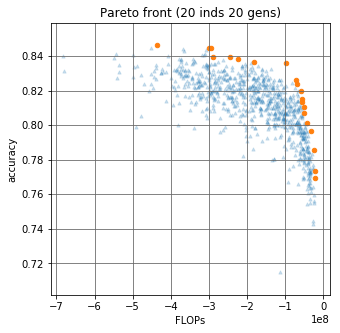} 
	\end{subfigure}
	\begin{subfigure}[b]{0.24\textwidth}   
		\centering 
		\includegraphics[width=\textwidth]{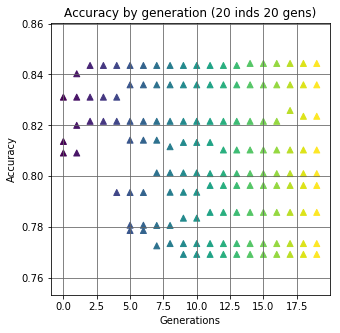}
	\end{subfigure}
	\begin{subfigure}[b]{0.24\textwidth}   
		\centering 
		\includegraphics[width=\textwidth]{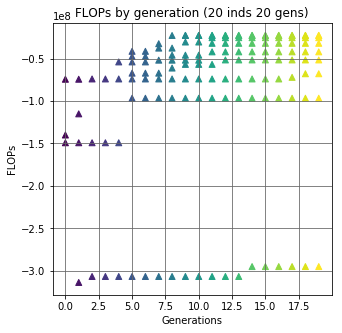}
	\end{subfigure}
	\begin{subfigure}[b]{0.24\textwidth}  
		\centering 
		\includegraphics[width=\textwidth]{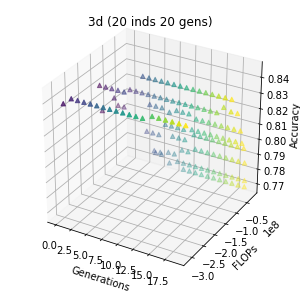}  
	\end{subfigure}
	\caption[]
	{\small 20 individuals and 20 generations} 
	\label{fig:plot_20gens_20inds}
\end{figure*}

\begin{figure*}
	\centering
	\begin{subfigure}[b]{0.24\textwidth}
		\centering
		\includegraphics[width=\textwidth]{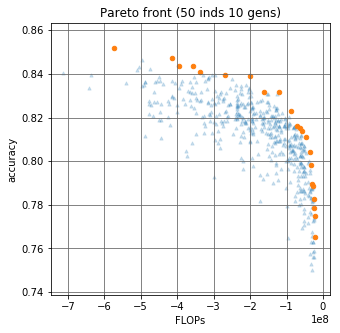}
	\end{subfigure}
	\begin{subfigure}[b]{0.24\textwidth}   
		\centering 
		\includegraphics[width=\textwidth]{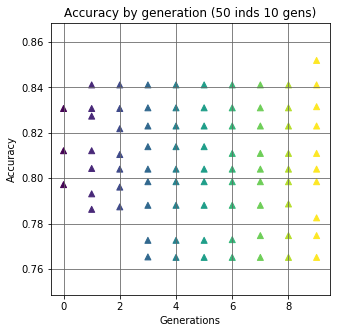}
	\end{subfigure}
	\begin{subfigure}[b]{0.24\textwidth}   
		\centering 
		\includegraphics[width=\textwidth]{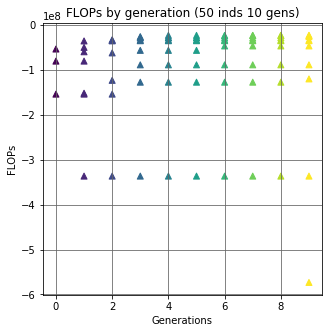}
	\end{subfigure}
	\begin{subfigure}[b]{0.24\textwidth}  
		\centering 
		\includegraphics[width=\textwidth]{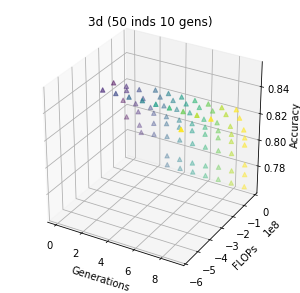}
	\end{subfigure}
	\caption[]
	{\small 50 individuals and 10 generations} 
	\label{fig:plot_50gens_50inds}
\end{figure*}

Fig. \ref{fig:plot_20gens_20inds} and Fig. \ref{fig:plot_50gens_50inds} show the experimental results of \textit{EXP-20-20} and \textit{EXP-50-10}, respectively, each of which is composed of four sub-figures. From the left to the right, the first sub-figure contains all of the solutions evaluated through the evolutionary process, where the x-axis represents the negative value of the FLOPs and y-axis shows the accuracy. The non-dominant solutions based on $\epsilon$-Pareto dominance \cite{EpsilonPareto_Laumanns} are in orange colour and the blue points indicate the others; The second sub-figure illustrates the evolutionary progress of the accuracy of non-dominant solutions based on $\epsilon$-Pareto dominance by each generation with the generation as x-axis and the classification accuracy as y-axis; The third sub-figure shows the changes of FLOPs of non-dominant solutions based on $\epsilon$-Pareto dominance during the evolutionary process, where the negative value of FLOPs is drawn toward the vertical axis and the generation is plotted toward the horizontal axis; The fourth sub-figure is generated by combining the second and third sub-figures into a 3D figure with x-axis, y-axis and z-axis represents the generation, the negative FLOPs value and the classification accuracy, respectively. The level of transparency reflects the depth in the 3D figure, i.e. the negative value of FLOPs carried by the point with less transparency is smaller than that represented by the more transparent point.

It can be observed that the negative value of FLOPs is plotted in the figure instead of the positive value, which is because that by using the negative value of FLOPs, it converts the optimization of this objective to a maximization problem in order to make it consistent to the other objective of maximizing the classification accuracy. After the conversion, the two objectives have the same optimization direction, which is easier to be understood and analysed. By looking into the first sub-figure of Fig. \ref{fig:plot_20gens_20inds} and Fig. \ref{fig:plot_50gens_50inds}, the non-dominant solutions achieved by both the experiments have formed a clear curve, which defines the Pareto front. When further investigating the Pareto front, it can be found that the two objectives contradict each other at some stage, i.e. the classification accuracy cannot be improved without increasing the FLOPs reflecting the complexity of CNNs, which means the problem of optimizing the two objectives of the classification accuracy and the FLOPs is an obvious multi-objective optimization problem. By comparing the Pareto fronts of the two experiments, especially the points with the lowest FLOPs and the highest accuracy, it can be learnt that \textit{EXP-50-10} provides more diverse non-dominant solutions, which also means the coverage of the non-dominant solutions of \textit{EXP-50-10} is larger than that of \textit{EXP-20-20}, even though the maximum generation of \textit{EXP-50-10} is only half of the generation of \textit{EXP-20-20}, so the larger population size in the proposed MOCNN method tends to produce more diverse non-dominant solutions, which therefore, provides more options for industrial users to choose. 

In regard with the convergence analysis, the second and third sub-figures can be utilized to analyse the convergence of the classification accuracy and FLOPs, respectively, and the third sub-figure presents an overview of the convergence of both of the objectives. Firstly, \textit{EXP-20-20} can be considered to be converged for both of the objectives. The classification accuracy changes a lot during the first 7 generations of evolution, and starts to fluctuate a bit until the end of the evolutionary process. As after the $12th$ generation, only very few non-dominant solutions shift a little bit, so \textit{EXP-20-20} can be deemed converged in terms of the classification accuracy. As shown in the third sub-figure of Fig. \ref{fig:plot_20gens_20inds}, the number of non-dominant solutions grows fast and the value of FLOPs quickly spreads to both directions before the $8th$ generation, but it is stabilizing until the $14th$ generation, after which the FLOPs hardly shift. Therefore, the FLOPs of \textit{EXP-20-20} is converged as well. The convergence progress of both objectives can be noticed in the fourth figure of Fig. \ref{fig:plot_20gens_20inds}. Secondly, with regard to the convergence of \textit{EXP-50-10}, it can be found that \textit{EXP-50-10} may need a lot more generations to converge. From the second sub-figure, there are obvious changes at the $1st$, $3rd$ and $10th$ generations, and between those generations, the shifts rarely happen, which indicates that the convergence speed of the experiment with 50 individuals is much slower and it needs more generations to converge in terms of the classification accuracy. For the FLOPs, the same pattern can be found as well, which is that at the $1st$ and $10th$ generations, the changes of non-dominant solutions are clearly seen, and rare changes take place for the other generations, so the objective of FLOPs also needs more time to converge. Therefore, \textit{EXP-50-10} hasn't reached the convergence, which can also be observed in the 3D sub-figure of Fig. \ref{fig:plot_50gens_50inds}. To summarize, the experiment with 20 individuals converges faster than that with 50 individuals, but the experiment with 50 individuals tend to provide more non-dominant solutions, which gains more coverage of the potential solutions. 

\subsection{MOCNN vs DenseNet-121}\label{sec:EPResults_DenseNet}

As described in Section \ref{sec:ParticleEncodingStrategy}, DenseNet-121 was set as the maximum complexity of the optimized CNNs, so DenseNet-121 is set as a benchmark, which is used as a comparison to the optimized non-dominant solution that has the best accuracy. As the classification accuracy of DenseNet-121 on CIFAR-10 was not reported in their paper, DenseNet-121 needs to be evaluated and compared with the optimized MOCNN. The same training process and the common-used data augmentation specified in \cite{DenseNet_Huang} are adopted to train both DenseNet-121 and the optimized MOCNN. The classification accuracy of DenseNet-121 is \%94.77 and the classification accuracy for the optimized MOCNN is \%95.51, which shows that the optimized MOCNN outperforms DenseNet-121 on CIFAR-10 dataset in terms of both the classification accuracy and the computational cost. The classification accuracies of DenseNet(k=12) of 40 layers (DenseNet40) and DenseNet(k=12) of 100 layers (DenseNet100\_12) are reported in \cite{DenseNet_Huang}, which are \%94.76 and \%95.90, respectively. The optimized MOCNN performs better than (DenseNet40), while a bit worse than (DenseNet100\_12). However, (DenseNet100\_12) is beyond the search space because it is more complex than DenseNet-121. Therefore, the optimized MOCNN has achieved a promising result among the DenseNets within the search space, and it may possibly outperform (DenseNet100\_12) if the search space is extended to include (DenseNet100\_12). 


\subsection{Computational Cost}\label{sec:EPResults_Computation}

As described in Section \ref{sec:FitnessEvaluation}, the CNNs represented by individuals are fully trained by Adam optimization, which consumes quite a large amount of computation. At the beginning, the experiment \textit{EXP-20-20} was tried on one GPU card, which took almost three weeks to finish the experiment, so a new infrastructure is proposed in Section \ref{sec:MocnnInfra} in order to leverage as many as GPU cards across multiple machines to dramatically reduce the running time. The experiment \textit{EXP-20-20} ran for about 3 days to finish the evolutionary process on 8 GPU cards, and the result of the experiment \textit{EXP-50-10} was achieved by running the program on 10 GPU cards for 3 days as well. It can be concluded the running time of the proposed MOCNN has been significantly plunged by utilizing as many available GPU cards as possible. 



\section{Conclusions}\label{sec:Conclusion}

This paper proposed a multi-objective EC method called MOCNN to search for the non-dominant solutions at the Pareto front by optimizing the two objectives of both the classification accuracy and the FLOPs reflecting the computational cost. The proposed MOCNN was designed and developed by designing a new encoding strategy to encode CNNs, choosing the two objectives that are critical to measuring the performance of CNNs, and applying a multi-objective particle swarm optimization algorithm called OMOPSO. Furthermore, an infrastructure is designed to boost the running speed of the proposed MOCNN, which can concurrently evaluate the CNNs on multiple GPU cards across multiple machines, and a Python library has been developed and released publicly. As non-dominant solutions are generated by the proposed MOCNN can be provided to the industrial users for them to choose one that suits their usage best, the overall goal of streamlining the usage of the state-of-the-art CNNs for image classification has been achieved. 

In terms of future works, there are several areas that we would like to explore. First of all, as this paper only explored the multi-objective optimization problem based on DenseNet structure, and there are more and more advanced CNN architectures \cite{SEBlock_Hu} \cite{CondenseNet_Huang} \cite{Polynet_Zhang} \cite{DualPathNet_Chen} invented which have achieved competitive or even better performance than DenseNet, it would be great to develop an algorithm that can effectively streamline the usage of the state-of-the-art CNNs, which can provide the potential non-dominant solutions without the constraint of one specific CNN structure; In addition, as this paper only explores the CNNs which are less complex than DenseNet-121 due to our hardware limitation, it would be more convincing to expand the search space with more complexity; Last but not least, although the FLOPs can reflect the computational cost, the inference time is not given for specific values of FLOPs, so it would be excellent to train a machine learning model to predict the inference time based on the value of FLOPs.

\newpage

\bibliographystyle{ACM-Reference-Format}
\bibliography{mocnn} 


\begin{thebibliography}{23}


\ifx \showCODEN    \undefined \def \showCODEN     #1{\unskip}     \fi
\ifx \showDOI      \undefined \def \showDOI       #1{#1}\fi
\ifx \showISBNx    \undefined \def \showISBNx     #1{\unskip}     \fi
\ifx \showISBNxiii \undefined \def \showISBNxiii  #1{\unskip}     \fi
\ifx \showISSN     \undefined \def \showISSN      #1{\unskip}     \fi
\ifx \showLCCN     \undefined \def \showLCCN      #1{\unskip}     \fi
\ifx \shownote     \undefined \def \shownote      #1{#1}          \fi
\ifx \showarticletitle \undefined \def \showarticletitle #1{#1}   \fi
\ifx \showURL      \undefined \def \showURL       {\relax}        \fi
\providecommand\bibfield[2]{#2}
\providecommand\bibinfo[2]{#2}
\providecommand\natexlab[1]{#1}
\providecommand\showeprint[2][]{arXiv:#2}

\bibitem[\protect\citeauthoryear{Bottou}{Bottou}{2010}]%
        {SGD_Bottou}
\bibfield{author}{\bibinfo{person}{L{\'e}on Bottou}.}
  \bibinfo{year}{2010}\natexlab{}.
\newblock \showarticletitle{Large-scale machine learning with stochastic
  gradient descent}.
\newblock In \bibinfo{booktitle}{{\em Proceedings of COMPSTAT'2010}}.
  \bibinfo{publisher}{Springer}, \bibinfo{pages}{177--186}.
\newblock


\bibitem[\protect\citeauthoryear{Chen, Li, Xiao, Jin, Yan, and Feng}{Chen
  et~al\mbox{.}}{2017}]%
        {DualPathNet_Chen}
\bibfield{author}{\bibinfo{person}{Yunpeng Chen}, \bibinfo{person}{Jianan Li},
  \bibinfo{person}{Huaxin Xiao}, \bibinfo{person}{Xiaojie Jin},
  \bibinfo{person}{Shuicheng Yan}, {and} \bibinfo{person}{Jiashi Feng}.}
  \bibinfo{year}{2017}\natexlab{}.
\newblock \showarticletitle{Dual path networks}. In \bibinfo{booktitle}{{\em
  Advances in Neural Information Processing Systems}}.
  \bibinfo{pages}{4467--4475}.
\newblock


\bibitem[\protect\citeauthoryear{Deb, Pratap, Agarwal, and Meyarivan}{Deb
  et~al\mbox{.}}{2002}]%
        {NSGA2_Deb}
\bibfield{author}{\bibinfo{person}{Kalyanmoy Deb}, \bibinfo{person}{Amrit
  Pratap}, \bibinfo{person}{Sameer Agarwal}, {and} \bibinfo{person}{TAMT
  Meyarivan}.} \bibinfo{year}{2002}\natexlab{}.
\newblock \showarticletitle{A fast and elitist multiobjective genetic
  algorithm: NSGA-II}.
\newblock \bibinfo{journal}{{\em IEEE transactions on evolutionary
  computation\/}} \bibinfo{volume}{6}, \bibinfo{number}{2}
  (\bibinfo{year}{2002}), \bibinfo{pages}{182--197}.
\newblock


\bibitem[\protect\citeauthoryear{Elsken, Metzen, and Hutter}{Elsken
  et~al\mbox{.}}{2018}]%
        {NAS_Elsken}
\bibfield{author}{\bibinfo{person}{Thomas Elsken}, \bibinfo{person}{Jan~Hendrik
  Metzen}, {and} \bibinfo{person}{Frank Hutter}.}
  \bibinfo{year}{2018}\natexlab{}.
\newblock \showarticletitle{Neural architecture search: A survey}.
\newblock \bibinfo{journal}{{\em arXiv preprint arXiv:1808.05377\/}}
  (\bibinfo{year}{2018}).
\newblock


\bibitem[\protect\citeauthoryear{Glorot, Bordes, and Bengio}{Glorot
  et~al\mbox{.}}{2011}]%
        {ReLU_Glorot}
\bibfield{author}{\bibinfo{person}{Xavier Glorot}, \bibinfo{person}{Antoine
  Bordes}, {and} \bibinfo{person}{Yoshua Bengio}.}
  \bibinfo{year}{2011}\natexlab{}.
\newblock \showarticletitle{Deep sparse rectifier neural networks}. In
  \bibinfo{booktitle}{{\em Proceedings of the fourteenth international
  conference on artificial intelligence and statistics}}.
  \bibinfo{pages}{315--323}.
\newblock


\bibitem[\protect\citeauthoryear{He, Zhang, Ren, and Sun}{He
  et~al\mbox{.}}{2015}]%
        {ResNet_He}
\bibfield{author}{\bibinfo{person}{Kaiming He}, \bibinfo{person}{Xiangyu
  Zhang}, \bibinfo{person}{Shaoqing Ren}, {and} \bibinfo{person}{Jian Sun}.}
  \bibinfo{year}{2015}\natexlab{}.
\newblock \showarticletitle{Deep Residual Learning for Image Recognition}.
\newblock \bibinfo{journal}{{\em CoRR\/}}  \bibinfo{volume}{abs/1512.03385}
  (\bibinfo{year}{2015}).
\newblock
\showeprint[arxiv]{1512.03385}
\showURL{%
\url{http://arxiv.org/abs/1512.03385}}


\bibitem[\protect\citeauthoryear{Hu, Shen, and Sun}{Hu et~al\mbox{.}}{2018}]%
        {SEBlock_Hu}
\bibfield{author}{\bibinfo{person}{Jie Hu}, \bibinfo{person}{Li Shen}, {and}
  \bibinfo{person}{Gang Sun}.} \bibinfo{year}{2018}\natexlab{}.
\newblock \showarticletitle{Squeeze-and-excitation networks}. In
  \bibinfo{booktitle}{{\em Proceedings of the IEEE Conference on Computer
  Vision and Pattern Recognition}}. \bibinfo{pages}{7132--7141}.
\newblock


\bibitem[\protect\citeauthoryear{Huang, Liu, Van~der Maaten, and
  Weinberger}{Huang et~al\mbox{.}}{2018}]%
        {CondenseNet_Huang}
\bibfield{author}{\bibinfo{person}{Gao Huang}, \bibinfo{person}{Shichen Liu},
  \bibinfo{person}{Laurens Van~der Maaten}, {and} \bibinfo{person}{Kilian~Q
  Weinberger}.} \bibinfo{year}{2018}\natexlab{}.
\newblock \showarticletitle{Condensenet: An efficient densenet using learned
  group convolutions}. In \bibinfo{booktitle}{{\em Proceedings of the IEEE
  Conference on Computer Vision and Pattern Recognition}}.
  \bibinfo{pages}{2752--2761}.
\newblock


\bibitem[\protect\citeauthoryear{Huang, Liu, and Weinberger}{Huang
  et~al\mbox{.}}{2016}]%
        {DenseNet_Huang}
\bibfield{author}{\bibinfo{person}{Gao Huang}, \bibinfo{person}{Zhuang Liu},
  {and} \bibinfo{person}{Kilian~Q. Weinberger}.}
  \bibinfo{year}{2016}\natexlab{}.
\newblock \showarticletitle{Densely Connected Convolutional Networks}.
\newblock \bibinfo{journal}{{\em CoRR\/}}  \bibinfo{volume}{abs/1608.06993}
  (\bibinfo{year}{2016}).
\newblock
\showeprint[arxiv]{1608.06993}
\showURL{%
\url{http://arxiv.org/abs/1608.06993}}


\bibitem[\protect\citeauthoryear{Ioffe and Szegedy}{Ioffe and Szegedy}{2015}]%
        {BN_Ioffe}
\bibfield{author}{\bibinfo{person}{Sergey Ioffe} {and}
  \bibinfo{person}{Christian Szegedy}.} \bibinfo{year}{2015}\natexlab{}.
\newblock \showarticletitle{Batch normalization: Accelerating deep network
  training by reducing internal covariate shift}.
\newblock \bibinfo{journal}{{\em arXiv preprint arXiv:1502.03167\/}}
  (\bibinfo{year}{2015}).
\newblock


\bibitem[\protect\citeauthoryear{Kingma and Ba}{Kingma and Ba}{2014}]%
        {Adam_Kingma}
\bibfield{author}{\bibinfo{person}{Diederik~P Kingma} {and}
  \bibinfo{person}{Jimmy Ba}.} \bibinfo{year}{2014}\natexlab{}.
\newblock \showarticletitle{Adam: A method for stochastic optimization}.
\newblock \bibinfo{journal}{{\em arXiv preprint arXiv:1412.6980\/}}
  (\bibinfo{year}{2014}).
\newblock


\bibitem[\protect\citeauthoryear{Krizhevsky and Hinton}{Krizhevsky and
  Hinton}{2009}]%
        {CIFAR10_Alex}
\bibfield{author}{\bibinfo{person}{Alex Krizhevsky} {and}
  \bibinfo{person}{Geoffrey Hinton}.} \bibinfo{year}{2009}\natexlab{}.
\newblock \bibinfo{booktitle}{{\em Learning multiple layers of features from
  tiny images}}.
\newblock \bibinfo{type}{{T}echnical {R}eport}.
  \bibinfo{institution}{Citeseer}.
\newblock


\bibitem[\protect\citeauthoryear{Laumanns, Thiele, Deb, and Zitzler}{Laumanns
  et~al\mbox{.}}{2002}]%
        {EpsilonPareto_Laumanns}
\bibfield{author}{\bibinfo{person}{Marco Laumanns}, \bibinfo{person}{Lothar
  Thiele}, \bibinfo{person}{Kalyanmoy Deb}, {and} \bibinfo{person}{Eckart
  Zitzler}.} \bibinfo{year}{2002}\natexlab{}.
\newblock \showarticletitle{Combining convergence and diversity in evolutionary
  multiobjective optimization}.
\newblock \bibinfo{journal}{{\em Evolutionary computation\/}}
  \bibinfo{volume}{10}, \bibinfo{number}{3} (\bibinfo{year}{2002}),
  \bibinfo{pages}{263--282}.
\newblock


\bibitem[\protect\citeauthoryear{Li}{Li}{2003}]%
        {NSPSO_Li}
\bibfield{author}{\bibinfo{person}{Xiaodong Li}.}
  \bibinfo{year}{2003}\natexlab{}.
\newblock \showarticletitle{A non-dominated sorting particle swarm optimizer
  for multiobjective optimization}. In \bibinfo{booktitle}{{\em Genetic and
  Evolutionary Computation Conference}}. Springer, \bibinfo{pages}{37--48}.
\newblock


\bibitem[\protect\citeauthoryear{Real, Moore, Selle, Saxena, Suematsu, Le, and
  Kurakin}{Real et~al\mbox{.}}{2017a}]%
        {EvolutionCNN_Moore}
\bibfield{author}{\bibinfo{person}{Esteban Real}, \bibinfo{person}{Sherry
  Moore}, \bibinfo{person}{Andrew Selle}, \bibinfo{person}{Saurabh Saxena},
  \bibinfo{person}{Yutaka~Leon Suematsu}, \bibinfo{person}{Quoc~V. Le}, {and}
  \bibinfo{person}{Alex Kurakin}.} \bibinfo{year}{2017}\natexlab{a}.
\newblock \showarticletitle{Large-Scale Evolution of Image Classifiers}.
\newblock \bibinfo{journal}{{\em CoRR\/}}  \bibinfo{volume}{abs/1703.01041}
  (\bibinfo{year}{2017}).
\newblock
\showeprint[arxiv]{1703.01041}
\showURL{%
\url{http://arxiv.org/abs/1703.01041}}


\bibitem[\protect\citeauthoryear{Real, Moore, Selle, Saxena, Suematsu, Tan, Le,
  and Kurakin}{Real et~al\mbox{.}}{2017b}]%
        {EvolutionNAS_Real}
\bibfield{author}{\bibinfo{person}{Esteban Real}, \bibinfo{person}{Sherry
  Moore}, \bibinfo{person}{Andrew Selle}, \bibinfo{person}{Saurabh Saxena},
  \bibinfo{person}{Yutaka~Leon Suematsu}, \bibinfo{person}{Jie Tan},
  \bibinfo{person}{Quoc Le}, {and} \bibinfo{person}{Alex Kurakin}.}
  \bibinfo{year}{2017}\natexlab{b}.
\newblock \showarticletitle{Large-scale evolution of image classifiers}.
\newblock \bibinfo{journal}{{\em arXiv preprint arXiv:1703.01041\/}}
  (\bibinfo{year}{2017}).
\newblock


\bibitem[\protect\citeauthoryear{Sierra and Coello}{Sierra and Coello}{2005}]%
        {OMOPSO_Sierra}
\bibfield{author}{\bibinfo{person}{Margarita~Reyes Sierra} {and}
  \bibinfo{person}{Carlos A~Coello Coello}.} \bibinfo{year}{2005}\natexlab{}.
\newblock \showarticletitle{Improving PSO-based multi-objective optimization
  using crowding, mutation and∈-dominance}. In \bibinfo{booktitle}{{\em
  International Conference on Evolutionary Multi-Criterion Optimization}}.
  Springer, \bibinfo{pages}{505--519}.
\newblock


\bibitem[\protect\citeauthoryear{Simonyan and Zisserman}{Simonyan and
  Zisserman}{2014}]%
        {VGG_Simonyan}
\bibfield{author}{\bibinfo{person}{Karen Simonyan} {and}
  \bibinfo{person}{Andrew Zisserman}.} \bibinfo{year}{2014}\natexlab{}.
\newblock \showarticletitle{Very Deep Convolutional Networks for Large-Scale
  Image Recognition}.
\newblock \bibinfo{journal}{{\em CoRR\/}}  \bibinfo{volume}{abs/1409.1556}
  (\bibinfo{year}{2014}).
\newblock
\showeprint[arxiv]{1409.1556}
\showURL{%
\url{http://arxiv.org/abs/1409.1556}}


\bibitem[\protect\citeauthoryear{Wang, Sun, Xue, and Zhang}{Wang
  et~al\mbox{.}}{2018a}]%
        {PSOCNN_Wang}
\bibfield{author}{\bibinfo{person}{B. Wang}, \bibinfo{person}{Y. Sun},
  \bibinfo{person}{B. Xue}, {and} \bibinfo{person}{M. Zhang}.}
  \bibinfo{year}{2018}\natexlab{a}.
\newblock \showarticletitle{Evolving Deep Convolutional Neural Networks by
  Variable-Length Particle Swarm Optimization for Image Classification}. In
  \bibinfo{booktitle}{{\em 2018 IEEE Congress on Evolutionary Computation
  (CEC)}}. \bibinfo{pages}{1--8}.
\newblock
\showDOI{%
\url{https://doi.org/10.1109/CEC.2018.8477735}}


\bibitem[\protect\citeauthoryear{Wang, Sun, Xue, and Zhang}{Wang
  et~al\mbox{.}}{2018b}]%
        {DECNN_Wang}
\bibfield{author}{\bibinfo{person}{Bin Wang}, \bibinfo{person}{Yanan Sun},
  \bibinfo{person}{Bing Xue}, {and} \bibinfo{person}{Mengjie Zhang}.}
  \bibinfo{year}{2018}\natexlab{b}.
\newblock \showarticletitle{A Hybrid Differential Evolution Approach to
  Designing Deep Convolutional Neural Networks for Image Classification}. In
  \bibinfo{booktitle}{{\em Australasian Joint Conference on Artificial
  Intelligence}}. Springer, \bibinfo{pages}{237--250}.
\newblock


\bibitem[\protect\citeauthoryear{Xie and Yuille}{Xie and Yuille}{2017}]%
        {GeneticCNN_Xie}
\bibfield{author}{\bibinfo{person}{L. Xie} {and} \bibinfo{person}{A. Yuille}.}
  \bibinfo{year}{2017}\natexlab{}.
\newblock \showarticletitle{Genetic CNN}. In \bibinfo{booktitle}{{\em 2017 IEEE
  International Conference on Computer Vision (ICCV)}}.
  \bibinfo{pages}{1388--1397}.
\newblock
\showISSN{2380-7504}
\showDOI{%
\url{https://doi.org/10.1109/ICCV.2017.154}}


\bibitem[\protect\citeauthoryear{Zhang, Li, Change~Loy, and Lin}{Zhang
  et~al\mbox{.}}{2017}]%
        {Polynet_Zhang}
\bibfield{author}{\bibinfo{person}{Xingcheng Zhang}, \bibinfo{person}{Zhizhong
  Li}, \bibinfo{person}{Chen Change~Loy}, {and} \bibinfo{person}{Dahua Lin}.}
  \bibinfo{year}{2017}\natexlab{}.
\newblock \showarticletitle{Polynet: A pursuit of structural diversity in very
  deep networks}. In \bibinfo{booktitle}{{\em Proceedings of the IEEE
  Conference on Computer Vision and Pattern Recognition}}.
  \bibinfo{pages}{718--726}.
\newblock


\bibitem[\protect\citeauthoryear{Zoph and Le}{Zoph and Le}{2016}]%
        {NAS_Zoph}
\bibfield{author}{\bibinfo{person}{Barret Zoph} {and} \bibinfo{person}{Quoc~V
  Le}.} \bibinfo{year}{2016}\natexlab{}.
\newblock \showarticletitle{Neural architecture search with reinforcement
  learning}.
\newblock \bibinfo{journal}{{\em arXiv preprint arXiv:1611.01578\/}}
  (\bibinfo{year}{2016}).
\newblock


\end{thebibliography}

\end{document}